\documentclass[conference]{IEEEtran}
\IEEEoverridecommandlockouts
\usepackage{cite}
\usepackage{amsmath,amssymb,amsfonts}
\usepackage{algorithmic}
\usepackage{graphicx}
\usepackage{textcomp}
\usepackage{xcolor}
\usepackage{comment}
\usepackage{url}
\usepackage{booktabs}
\usepackage{multirow}
\def\BibTeX{{\rm B\kern-.05em{\sc i\kern-.025em b}\kern-.08em
    T\kern-.1667em\lower.7ex\hbox{E}\kern-.125emX}}

\usepackage[hidelinks]{hyperref}
\usepackage{makecell}
  
\begin{document}

\title{How Well Can LLMs Verbalize Mathematical Formulas?\\A Multi-Domain Evaluation}

\author{\IEEEauthorblockN{Peter Volbach, Tim Schlippe}
\IEEEauthorblockA{
\textit{IU International University of Applied Sciences}\\
Germany \\
tim.schlippe@iu.org}
}

\maketitle

\begin{abstract}
Large Language Models (LLMs) have shown strong capabilities across natural language tasks. However, their ability to generate accurate written verbalizations of mathematical formulas has not yet been systematically evaluated. Such verbalization is critical for text-to-speech systems serving audio-based learning and visually impaired users. Consequently, we systematically evaluated the formula verbalization performance of five state-of-the-art LLMs across six scientific domains. Since existing corpora do not jointly provide multi-domain coverage, validated transcription variants, textual context, and controlled formula complexity, we constructed \textit{Read-Aloud Math}, a balanced multi-domain corpus containing LaTeX formulas, manually validated concise \textit{simple} transcriptions and explanatory \textit{extended} transcriptions, multiple transcription variants, surrounding textual context, domain labels, and formula complexity levels. 
Based on multi-reference BLEU, GPT-5.2 achieved the highest score for \textit{simple} transcriptions (75.9\% without textual context, 64.9\% with textual context). In the context-free setting, however, its score did not differ significantly from those of Grok 4.1 Fast or Claude Sonnet 4.5. Claude Sonnet 4.5 obtained the highest score for \textit{extended} transcriptions without textual context (36.5\%), and Grok 4.1 Fast with textual context (48.8\%). For \textit{extended} transcriptions, \textit{Statistics} and \textit{Mathematics} scored highest across the two context settings, while \textit{Physics} scored lowest. The relationship between formula length and BLEU differed by transcription type: For \textit{simple} transcriptions, BLEU decreased significantly with increasing formula length without context, while the relationship was not significant with context. For \textit{extended} transcriptions, BLEU increased significantly with formula length in both context settings. Our notation-level analyses showed that parentheses and other grouping symbols were most frequently involved in transcription differences, while \textit{variable names}, \textit{capitalization}, and \textit{Greek letters} were associated with lower BLEU scores. These findings indicate that LLM-based formula verbalization is promising, but remains sensitive to transcription type, textual context, and characteristics of the underlying formulas, including domain, complexity, and notation.
\end{abstract}

\begin{IEEEkeywords}
formula verbalization, mathematical formulas, large language models, text-to-speech, accessibility
\end{IEEEkeywords}

\section{Introduction}

The growing use of audio-based learning materials, from audiobooks to podcast-style lecture content, has created new opportunities for flexible knowledge acquisition~\cite{Hew2009,Ahmad2022,Meden2024}. Students increasingly consume educational content while commuting, exercising, or performing activities where visual attention is unavailable~\cite{Crompton2018}. In parallel, text-to-speech (TTS) technology has become a standard interface for digital text access.

Beyond educational convenience, accessible mathematics is crucial for visually impaired individuals who rely on screen readers and TTS systems to access scientific material~\cite{Nazemi2012, Frankel2016}. For these users, accurate and unambiguous verbalization of mathematical formulas is essential (although this work does not target a specific accessibility standard).

Despite these advances, mathematical formulas remain a persistent challenge for TTS systems, particularly in science, technology, engineering, and mathematics (STEM) education. The core problem lies not in speech synthesis itself, but in determining \textit{what text} should be synthesized. Mathematical formulas require verbalization---the conversion of symbolic mathematical notation into natural language text---before speech rendering. Without this step, formulas may be omitted from audio versions or verbalized in ways that prevent listeners from reconstructing the original notation. Even advanced TTS systems often struggle with intelligibility when processing mathematical expressions~\cite{Roychowdhury2025}.

Despite decades of research on mathematical formula verbalization, the problem remains only partially solved, particularly when balancing comprehensibility, conciseness, and domain-specific conventions across diverse areas of mathematics and science. Traditional approaches to formula verbalization have relied primarily on rule-based systems \cite{Thinn2020, Bier2019, Dumkasem2019, Frankel2016, Bier2015, Wararat2013, Nazemi2012, Wongkia2012, Ferreira2004}. These systems apply deterministic transformation rules that map mathematical symbols and structures to predefined spoken verbalizations. While such approaches ensure structural consistency and reproducibility, they operate largely at the syntactic level and lack broader discourse awareness. As a result, they struggle with context-dependent interpretations and domain-specific conventions. For instance, $x'$ may be read as ``x prime'' or ``the derivative of x'', depending on context.

Recent advances in large language models (LLMs) have demonstrated strong capabilities in natural language understanding and generation, including context-sensitive interpretation and adaptation to domain-specific usage. This suggests that they may also be useful for context-aware mathematical verbalization \cite{Bier2025, Roychowdhury2025, Hyeon2025}. Unlike rule-based systems, LLMs can condition their outputs on surrounding text, enabling more flexible and semantically informed verbalization strategies.

Formula verbalization can serve two complementary goals: faithfully rendering the symbolic structure and supporting comprehension in audio-only settings through brief explanatory phrasing. Consequently, we distinguish between concise \textit{simple} transcriptions and explanatory \textit{extended} transcriptions, allowing us to evaluate both notation-oriented and comprehension-oriented verbalization.

Our work addresses this verbalization challenge: Given a mathematical formula in LaTeX format, what natural language text should be generated for a TTS system to speak? We focus on LaTeX since it remains the predominant format for authoring mathematical expressions in scientific and educational documents. While we do not perform speech synthesis itself, we evaluate how well LLMs can generate appropriate text verbalizations that would enable effective audio rendering of mathematical content. Specifically, we address the following three research questions: 

\begin{enumerate}
\item How accurately can state-of-the-art LLMs generate formula verbalizations from LaTeX representations?
\item What is the impact of textual context on formula verbalization performance?
\item How does formula verbalization performance vary across scientific domains, formula complexity levels, and notation features?
\end{enumerate}

Throughout this paper, we use \textit{verbalization} 
to refer to the general task of converting mathematical notation into spoken 
language, and \textit{transcription} to refer to the specific textual outputs 
generated by this process (\textit{simple} transcriptions and \textit{extended} transcriptions).

This paper makes two main contributions: First, we introduce \textit{Read-Aloud Math}, a balanced multi-domain corpus of mathematical formulas from six scientific domains (\textit{Mathematics}, \textit{Statistics}, \textit{Physics}, \textit{Chemistry}, \textit{Economics}, and \textit{Computer Science}), featuring multiple manually validated variants of \textit{simple} transcriptions and \textit{extended} transcriptions and surrounding textual context from Wikipedia articles, and make the corpus publicly available\footnote{\url{https://github.com/peter1706/verbalize-math-corpus}}. Second, we present a comprehensive evaluation of five state-of-the-art LLMs across domains, formula complexity levels, notation features, and textual context.

The remainder of this paper is structured as follows: Section~\ref{sec:Related Work} reviews related work. Section~\ref{sec:setup} describes our corpus construction methodology and the experimental setup for LLM evaluation. Section~\ref{sec:results} presents our evaluation results, and Section~\ref{sec:conclusion} concludes and indicates future research directions.

\section{Related Work}
\label{sec:Related Work}

Research on mathematical formula verbalization spans rule-based systems, statistical translation approaches, pre-trained sequence-to-sequence models, and more recently LLMs.

\subsection{Rule-Based Formula Verbalization}

Early work established that intelligible verbalization requires explicit modeling of hierarchical mathematical structure rather than linear symbol rendering \cite{Ferreira2004}. Rule-based systems such as MathSpeak, ClearSpeak, i-Math, and subsequent adaptable rule engines \cite{Nazemi2012,Wongkia2012,Wararat2013,Frankel2016,Bier2015,Bier2019,Mazzei2019,Dumkasem2019} provide deterministic and accessibility-oriented outputs. While these systems ensure reproducibility and structural transparency, they require extensive manual rule engineering and offer limited adaptability to discourse context or domain-specific variation.

Grammar-based approaches provide an alternative to traditional rule-based verbalization systems. The GF Mathematical Grammar Library, developed within the WebALT and MOLTO projects, supports multilingual mathematical verbalization from OpenMath representations using the Grammatical Framework (GF), providing deterministic semantic verbalization while requiring handcrafted grammars and semantic representations \cite{Strotmann2005,Caprotti2012}.

\subsection{Machine Translation-Based Formula Verbalization}

To reduce handcrafted rule design, formula verbalization has been framed as a translation problem. Statistical and neural approaches learn LaTeX–spoken mappings from parallel corpora \cite{Thinn2020, Dong2022}. Transfer learning further advanced this paradigm: MathReader \cite{Hyeon2025} employs a T5-based sequence-to-sequence model fine-tuned on MathBridge \cite{Jung2024}. Related work extends this approach to pipelines converting formula images into LaTeX and spoken-English transcriptions, comparing fine-tuned sequence-to-sequence models such as BART and T5 with GPT-based LLMs \cite{Feichter2024}. Sequence-to-sequence models achieve lower error rates for the LaTeX-to-text step. Overall, treating formula verbalization as a machine translation task improves fluency but remains dependent on training data coverage and domain scope.

\subsection{LLM-Based Formula Verbalization}

Recent studies suggest that state-of-the-art LLMs possess contextual reasoning capabilities relevant to mathematical language. Newer GPT-based systems generate plausible formula verbalizations \cite{Bier2025}, and LLMs resolve context-dependent identifier ambiguities in scientific documents \cite{Dev2024}. However, inconsistency and reproducibility remain concerns \cite{Bier2025}. Downstream evaluations of mathematical text-to-speech systems further demonstrate that intelligibility varies substantially across expression categories \cite{Roychowdhury2025}, emphasizing the importance of high-quality textual verbalization prior to speech synthesis.

\subsection{Corpora for Formula Verbalization}

Since existing corpora differ in scale, domain coverage, annotation depth, and spoken alignment, they restrict which aspects of formula verbalization can be evaluated. Large-scale web corpora such as OpenWebMath \cite{Paster2023} and InfiMM-WebMath-40B \cite{Han2024} preserve LaTeX formulas in natural-language context at web scale but lack spoken alignments. Tex80M \cite{Li2025} focuses on visual recognition of mathematical expressions. Thinn et al.~\cite{Thinn2020} constructed a LaTeX–spoken mathematics corpus from textbooks, but it is limited to mathematics, does not incorporate context, and does not report structural complexity characteristics. Formula2Text \cite{Feichter2024} provides formula images with corresponding LaTeX code and multiple spoken transcriptions, but targets image-based transcription rather than controlled multi-domain verbalization evaluation.  MathBridge \cite{Jung2024} provides large-scale LaTeX-spoken English pairs and a limited surrounding-text window, but models only a single automatically generated verbalization per formula. S2L \cite{Korzh2024} introduces multilingual spoken mathematical data but targets audio-to-LaTeX recognition rather than controlled text-based verbalization generation. Earlier arXiv-based corpora such as the KDD Cup 2003 dataset \cite{Gehrke2003} demonstrated the feasibility of large-scale structured LaTeX collections, yet were domain-specific and not designed for verbalization research.

\subsection{Research Gaps}

Prior work shows that intelligible formula verbalization requires preserving hierarchical structure and resolving notation-dependent ambiguities \cite{Ferreira2004, Nazemi2012}. Rule-based and grammar-based systems provide deterministic verbalizations but require handcrafted rules or grammars, limiting scalability and contextual adaptability \cite{Bier2015,Bier2019,Strotmann2005,Caprotti2012}. Data-driven and neural approaches \cite{Thinn2020, Dong2022, Hyeon2025}, including recent LLM-based methods \cite{Bier2025, Dev2024}, capture contextual dependencies and generate fluent outputs. However, several gaps remain:

\begin{itemize}
    \item \textbf{Multi-Level Transcription.} Existing systems produce either minimal readings or explanatory outputs, without systematically distinguishing \textit{simple} and \textit{extended} transcriptions.
    \item \textbf{Multi-Domain Evaluation.} Evaluations are typically restricted to single domains \cite{Hyeon2025}, despite substantial variation in notation across \textit{Mathematics}, \textit{Physics}, \textit{Statistics}, \textit{Chemistry}, \textit{Economics}, and \textit{Computer Science}.
    \item \textbf{Transcription Variant Coverage.} Large-scale corpora such as MathBridge \cite{Jung2024} provide only one canonical transcription per formula, although mathematical formulas allow multiple legitimate transcriptions.
    \item \textbf{Context-Aware LLM Evaluation.} Systematic comparison of verbalization quality with and without surrounding document context across multiple LLMs is missing.
    \item \textbf{Complexity Stratification.} Most corpora do not systematically vary formula complexity, making it difficult to analyze how performance changes across complexity levels.
\end{itemize}

These limitations motivate our multi-domain evaluation of state-of-the-art LLMs, explicitly comparing \textit{context-free} and \textit{context-aware} generation, incorporating both transcription types, and analyzing performance across domains, formula complexity levels, and notation features.

\section{Experimental Setup}
\label{sec:setup}

This section describes the construction and annotation of our corpus, the evaluated LLMs and prompting settings, and the evaluation methodology. The experimental design systematically varies transcription type and textual context while controlling for scientific domain and formula complexity.

\subsection{Corpus Construction}

We constructed \textit{Read-Aloud Math}, a balanced multi-domain corpus of mathematical formulas with manually validated transcriptions for evaluating formula verbalization systems. Rather than maximizing scale, it was designed for controlled analysis across domains and formula complexity levels.

\subsubsection{Domain Selection and Data Source}

Table~\ref{tab:corpus_stats} summarizes our corpus of 120 formulas (20 per domain) across six scientific domains: \textit{Mathematics}, \textit{Statistics}, \textit{Physics}, \textit{Chemistry}, \textit{Economics}, and \textit{Computer Science}. These domains represent major STEM areas with distinct notational conventions. \textit{Mathematics} and \textit{Statistics} primarily use abstract symbolic notation, \textit{Physics} relies on Greek-letter quantities, \textit{Chemistry} employs subscripts and superscripts for molecular structures, \textit{Economics} features optimization notation, and \textit{Computer Science} includes algorithmic and complexity expressions.

Formulas were extracted from Wikipedia-derived documents in the OpenWebMath corpus \cite{Paster2023}, which preserves both LaTeX expressions and their surrounding article text. Restricting extraction to Wikipedia ensures publicly accessible, community-reviewed scientific content and coherent discourse.

To ensure systematic coverage of structural complexity, each domain includes five formulas with 1--10, 11--20, 21--30, and 31--40 symbols. Symbol count is obtained by parsing LaTeX with pylatexenc\footnote{\url{https://github.com/phfaist/pylatexenc}}, normalizing the parsed output by inserting whitespace around operators, and counting the resulting tokens (commands, operators, and identifiers each count as one). This stratified sampling enables controlled analysis of performance under increasing structural difficulty.

\begin{table}[t]
\centering
\caption{Corpus statistics by domain}
\label{tab:corpus_stats}
\setlength{\tabcolsep}{1.5pt}
\footnotesize
\begin{tabular}{lccc}
\hline
\textbf{Domain} & \textbf{\O Simple vars.} &  \textbf{\O Extended vars.} & \textbf{\O Context chars.}\\
\hline
Mathematics     & 3.1 & 4.9 & ~~9,065.7 \\
Statistics      & 3.6 & 4.7 & ~~9,787.4 \\
Physics         & 3.7 & 4.8 & 16,481.8 \\
Chemistry       & 4.2 & 5.0 & 15,306.2 \\
Economics       & 3.8 & 5.0 & ~~9,000.9 \\
Computer Science   & 3.8 & 5.0 & 11,635.6 \\
\hline
\textbf{Total}  & 3.7 & 4.9 & 11,879.6 \\
\hline
\end{tabular}
\end{table}

\subsubsection{Transcription Types and Variants}

Each formula includes two complementary transcription types. \textit{Simple} transcriptions provide concise, notation-oriented readings using standard mathematical terminology without explaining the meanings of individual symbols. For example, $E = mc^2$ becomes ``Capital E equals m c squared''. \textit{Extended} transcriptions preserve the formula structure while adding brief explanatory wording and symbol interpretations intended to improve comprehensibility in audio-only settings. The same formula becomes ``Energy E equals mass m times the speed of light c squared''. Extended transcriptions were limited to approximately 1.5--3 times the length of their simple counterparts to avoid unnecessarily verbose explanations.

For each formula, the final reference set contains up to five distinct variants per transcription type to capture natural transcription diversity and domain-specific preferences. These variants provide multiple reference transcriptions for evaluation and reflect legitimate variation in mathematical speech.

\subsubsection{Annotation Process}

For each formula, each of the five LLMs later evaluated in our experiments generated one initial transcription per transcription type: Gemini 2.5 Flash (Google), DeepSeek V3.2 (DeepSeek), GPT-5.2 (OpenAI), Claude Sonnet 4.5 (Anthropic), and Grok 4.1 Fast (xAI). Thus, five initial \textit{simple} and five initial \textit{extended} transcriptions were generated per formula. Each transcription was subsequently manually post-edited. Two expert annotators independently reviewed all outputs using shared annotation guidelines, which required each transcription to preserve all symbols and structural relations, avoid unsupported meaning, and remain natural for spoken rendering. Discrepancies were resolved through discussion and consensus, and the final transcriptions were checked for completeness, correctness, and naturalness.

Since different LLMs occasionally produced identical transcriptions and post-editing could also yield identical outputs, some formulas contain fewer than five distinct reference variants. This is reflected in the average number of variants per transcription type in Table~\ref{tab:corpus_stats}. The same OpenRouter-based request setup was used to generate both the initial reference transcriptions and the evaluated outputs, with no explicit generation parameters specified.

\subsection{Evaluated LLMs and Experimental Design}

We evaluated the five LLMs described above in February 2026. Although all drafts were independently reviewed, post-edited, and validated by two expert annotators, residual LLM-specific wording may increase lexical overlap between an LLM's output and the references, which we consider when interpreting BLEU-based comparisons.

Each LLM was evaluated in two settings: \textit{context-free} and \textit{context-aware}. In the \textit{context-free} setting, only the LaTeX formula was provided, whereas the \textit{context-aware} setting additionally included surrounding Wikipedia text. In the result tables, these settings are referred to as ``without textual context'' and ``with textual context'', respectively. For both settings, we requested \textit{simple} and \textit{extended} transcriptions. Each LLM generated one output per formula, transcription type, and context setting, with no repeated generations.

All LLMs were accessed through the OpenRouter Chat Completions API in February 2026 with the model identifiers \texttt{google/gemini-2.5-flash}, \texttt{deepseek/deepseek-v3.2}, \texttt{openai/gpt-5.2}, \texttt{anthropic/claude-sonnet-4.5}, and \texttt{x-ai/grok-4.1-fast}. Only the prompt was specified. Generation parameters such as temperature, top-p, maximum output length, and reasoning settings were left unspecified. OpenRouter automatically routed each request to a provider hosting the requested LLM, which applied its own defaults.

\subsection{Prompting Strategy}

We used four fixed prompt variants built from the same core instructions, one for each combination of transcription type (\textit{simple} or \textit{extended}) and context setting (\textit{context-free} or \textit{context-aware}). The complete verbatim prompts are provided in the public repository. All variants instructed the LLMs to verbalize the complete formula, produce clear and natural spoken English, express numbers as words, state capitalization explicitly, avoid LaTeX notation and mathematical symbols, and return only the transcription. They differed only through condition-specific additions: the context-aware variants additionally provided the text preceding and following the formula and instructed the LLM to use relevant contextual information, while the extended-transcription variants additionally requested a concise but more informative transcription with semantic clarification where appropriate. The context-aware extended variant combined both additions. All variants ended with the same self-check instruction asking the LLM to verify that all prompt instructions had been followed. No chain-of-thought or step-by-step reasoning was requested.

\subsection{Evaluation}

We evaluated transcription quality using BLEU computed with the sacreBLEU implementation \cite{Post2018} (default 4-gram setting). We selected multi-reference BLEU since mathematical formulas can have several valid transcriptions, and the use of multiple manually validated reference variants allows legitimate lexical variation to be captured while providing a standardized and reproducible basis for comparing LLM outputs. BLEU was computed separately for each formula and then averaged across formulas, enabling paired comparisons across LLMs. All available reference variants were provided simultaneously, and n-gram matches were counted against the maximum reference frequency across variants, following established machine translation practice. As BLEU measures n-gram overlap with the reference transcriptions, it does not directly assess semantic correctness or whether a listener could reconstruct the original formula. In addition, as the number of available reference variants differs slightly across formulas and domains, domain-level BLEU comparisons may also be influenced by differences in reference coverage.

To assess whether differences between LLMs were statistically reliable, we performed paired two-sided $t$-tests on formula-level BLEU scores. Since all LLMs were evaluated on the same 120 formulas, comparisons were paired by formula ID. Tests were conducted separately for each transcription type and setting. Within each combination, we performed four paired tests comparing the highest-scoring LLM with each of the other four LLMs. As multiple significance tests increase the chance of incorrectly identifying a difference as statistically significant, we applied the Holm correction to these four $p$-values at an overall significance level of $\alpha = 0.05$.

To assess the relationship between formula length and LLM performance beyond the four symbol-count ranges, we also used Spearman's rank correlation. We correlated each formula's symbol count with its BLEU score averaged across the five LLMs, separately for each transcription type and context setting. The Holm correction was applied to these four correlation tests at an overall significance level of $\alpha = 0.05$.

To better understand which formula characteristics were challenging for LLM-based formula verbalization, we additionally analyzed performance across notation features extracted from the LaTeX representation using rule-based pattern matching. Examples included fractions, subscripts, superscripts, Greek letters, and optimization operators. We first recorded how frequently these features occurred in the corpus and then computed average formula-level BLEU scores for formulas containing each feature. To complement this formula-level analysis, we extracted token-level confusion pairs from aligned reference and generated transcriptions. The resulting confusion pairs were grouped by notation feature and analyzed separately for each transcription type and setting.

\begin{table*}[t]
\centering
\caption{BLEU scores (\%) of LLM-generated formula transcriptions across domains.}
\label{tab:bleu_domain_full}
\setlength{\tabcolsep}{1pt}
\footnotesize
\begin{tabular}{l|ccccc|c|ccccc|c|ccccc|c|ccccc|c}
\toprule
& \multicolumn{12}{c}{\textbf{Simple transcription}}
& \multicolumn{12}{c}{\textbf{Extended transcription}} \\
\cmidrule(lr){2-13}\cmidrule(lr){14-25}
& \multicolumn{6}{c}{\textbf{Without textual context}}
& \multicolumn{6}{c}{\textbf{With textual context}}
& \multicolumn{6}{c}{\textbf{Without textual context}}
& \multicolumn{6}{c}{\textbf{With textual context}} \\
\cmidrule(lr){2-7}\cmidrule(lr){8-13}\cmidrule(lr){14-19}\cmidrule(lr){20-25}
\textbf{Domain}
& Gem & Deep & GPT & Clau & Gr & \textbf{Avg.}
& Gem & Deep & GPT & Clau & Gr & \textbf{Avg.}
& Gem & Deep & GPT & Clau & Gr & \textbf{Avg.}
& Gem & Deep & GPT & Clau & Gr & \textbf{Avg.} \\
\midrule
Mathematics & 59.9 & 63.0 & 68.7 & 67.4 & \textbf{76.3} & 67.0 & 43.9 & 60.2 & 57.5 & 55.3 & \textbf{60.6} & 55.5 & 30.6 & 23.4 & 20.3 & 38.2 & \textbf{39.4} & 30.4 & 36.7 & 41.4 & 45.1 & 50.1 & \textbf{54.4} & 45.6 \\
Statistics & 75.7 & 68.3 & 74.8 & 80.2 & \textbf{81.3} & 76.0 & 55.8 & 69.2 & 66.8 & 71.0 & \textbf{73.8} & 67.3 & 38.0 & 23.2 & 23.2 & 37.4 & \textbf{43.5} & 33.0 & 34.1 & 37.7 & 45.1 & 47.6 & \textbf{59.2} & 44.8 \\
Physics & 60.7 & 54.3 & \textbf{80.8} & 63.8 & 61.0 & 64.1 & 37.1 & 46.8 & \textbf{64.8} & 52.3 & 52.3 & 50.7 & 22.3 & 19.5 & 19.1 & 31.3 & \textbf{33.1} & 25.1 & 25.7 & 26.7 & 37.9 & \textbf{44.8} & 39.6 & 34.9 \\
Chemistry & 60.3 & 54.3 & \textbf{77.5} & 63.0 & 75.5 & 66.1 & 42.1 & 48.6 & \textbf{63.8} & 53.1 & 45.3 & 50.6 & 25.5 & 24.8 & 21.9 & 36.7 & \textbf{38.3} & 29.4 & 28.8 & 29.1 & \textbf{45.8} & 40.1 & 43.2 & 37.4 \\
Economics & 64.9 & 63.6 & 77.1 & \textbf{79.0} & 75.6 & 72.0 & 45.6 & 54.1 & 67.2 & \textbf{73.9} & 64.6 & 61.1 & 24.0 & 20.2 & 21.5 & \textbf{38.6} & 27.2 & 26.3 & 33.2 & 24.9 & 43.1 & 47.9 & \textbf{51.9} & 40.2 \\
Computer Science & 70.5 & 61.1 & \textbf{76.5} & 74.3 & 70.2 & 70.5 & 48.8 & 65.9 & \textbf{69.5} & 61.2 & 64.6 & 62.0 & 34.4 & 25.2 & 20.3 & \textbf{36.9} & 32.5 & 29.9 & 32.2 & 40.0 & 38.9 & 40.4 & \textbf{44.6} & 39.2 \\
\midrule
\textbf{Total} & 65.3 & 60.7 & \textbf{75.9} & 71.3 & 73.3 & 69.3 & 45.5 & 57.5 & \textbf{64.9} & 61.1 & 60.2 & 57.9 & 29.1 & 22.7 & 21.0 & \textbf{36.5} & 35.6 & 29.0 & 31.8 & 33.3 & 42.7 & 45.2 & \textbf{48.8} & 40.4 \\
\bottomrule
\end{tabular}
\end{table*}

\begin{table*}[t]
\centering
\caption{BLEU scores (\%) of LLM-generated formula transcriptions across formula symbol-count ranges.}
\label{tab:bleu_complexity}
\setlength{\tabcolsep}{1.5pt}
\footnotesize
\begin{tabular}{l|ccccc|c|ccccc|c|ccccc|c|ccccc|c}
\toprule
& \multicolumn{12}{c}{\textbf{Simple transcription}}
& \multicolumn{12}{c}{\textbf{Extended transcription}} \\
\cmidrule(lr){2-13}\cmidrule(lr){14-25}
& \multicolumn{6}{c}{\textbf{Without textual context}}
& \multicolumn{6}{c}{\textbf{With textual context}}
& \multicolumn{6}{c}{\textbf{Without textual context}}
& \multicolumn{6}{c}{\textbf{With textual context}} \\
\cmidrule(lr){2-7}\cmidrule(lr){8-13}\cmidrule(lr){14-19}\cmidrule(lr){20-25}
\textbf{\#symbols}
& Gem & Deep & GPT & Clau & Gr & \textbf{Avg.}
& Gem & Deep & GPT & Clau & Gr & \textbf{Avg.}
& Gem & Deep & GPT & Clau & Gr & \textbf{Avg.}
& Gem & Deep & GPT & Clau & Gr & \textbf{Avg.} \\
\midrule
1-10 & 76.4 & 74.9 & 84.0 & 80.4 & \textbf{85.2} & 80.2 & 43.2 & 72.0 & 64.7 & 54.7 & \textbf{74.4} & 61.8 & 14.3 & 14.1 & 12.1 & \textbf{23.4} & 23.1 & 17.4 & 29.2 & 28.5 & 40.5 & 34.2 & \textbf{45.3} & 35.5 \\
11-20 & 55.0 & 55.7 & \textbf{70.1} & 61.5 & 67.0 & 61.9 & 42.7 & 48.8 & \textbf{64.3} & 59.0 & 53.3 & 53.6 & 27.8 & 23.5 & 20.4 & \textbf{33.1} & 32.4 & 27.4 & 26.2 & 30.0 & 36.9 & 40.2 & \textbf{45.6} & 35.8 \\
21-30 & 67.8 & 61.1 & \textbf{81.7} & 75.4 & 70.7 & 71.4 & 52.0 & 57.9 & 69.6 & \textbf{71.2} & 59.6 & 62.1 & 39.0 & 27.8 & 28.8 & \textbf{47.1} & 46.8 & 37.9 & 35.7 & 41.1 & 49.0 & \textbf{56.6} & 52.3 & 46.9 \\
31-40 & 62.0 & 51.3 & 67.7 & 67.7 & \textbf{70.2} & 63.8 & 44.2 & 51.1 & \textbf{61.1} & 59.7 & 53.5 & 53.9 & 35.5 & 25.6 & 22.8 & \textbf{42.4} & 40.2 & 33.3 & 36.2 & 33.6 & 44.2 & 49.7 & \textbf{52.1} & 43.2 \\
\bottomrule
\end{tabular}
\end{table*}

\begin{table}[t]
\centering
\caption{Paired $t$-test results comparing the highest-scoring LLM with all other LLMs in each setting. Asterisks indicate significance after Holm correction ($\alpha=0.05$).}
\label{tab:significance}
\setlength{\tabcolsep}{2.5pt}
\footnotesize
\begin{tabular}{llccc}
\toprule
\textbf{Setting} & \textbf{Comparison} & \textbf{$\Delta$ BLEU (\%)} & \textbf{$p$} & \textbf{Sig.} \\
\midrule
\multirow{4}{*}{\makecell[l]{Simple transcription\\Without textual context}}
& GPT--Gemini & 10.57 & $<.001$ & * \\
& GPT--DeepSeek & 15.15 & $<.001$ & * \\
& GPT--Claude & ~4.62 & ~~~.099 &  \\
& GPT--Grok & ~2.60 & ~~~.339 &  \\
\midrule
\multirow{4}{*}{\makecell[l]{Simple transcription\\With textual context}}
& GPT--Gemini & 19.40 & $<.001$ & * \\
& GPT--DeepSeek & ~7.46 & ~~~.022 &  \\
& GPT--Claude & ~3.78 & ~~~.197 &  \\
& GPT--Grok & ~4.74 & ~~~.156 &  \\
\midrule
\multirow{4}{*}{\makecell[l]{Extended transcription\\Without textual context}}
& Claude--Gemini & ~7.35 & $<.001$ & * \\
& Claude--DeepSeek & 13.77 & $<.001$ & * \\
& Claude--GPT & 15.45 & $<.001$ & * \\
& Claude--Grok & ~0.85 & .610 &  \\
\midrule
\multirow{4}{*}{\makecell[l]{Extended transcription\\With textual context}}
& Grok--Gemini & 17.03 & $<.001$ & * \\
& Grok--DeepSeek & 15.51 & $<.001$ & * \\
& Grok--GPT & ~6.17 & ~~~.003 & * \\
& Grok--Claude & ~3.68 & ~~~.143 &  \\
\bottomrule
\end{tabular}
\end{table}

\begin{table}[t]
\centering
\caption{Notation coverage of the formula corpus.}
\label{tab:notation_coverage}
\footnotesize
\setlength{\tabcolsep}{2.2pt}
\begin{tabular}{llrrl}
\toprule
\textbf{Category} & \textbf{Feature} & \textbf{Count} & \textbf{\%} & \textbf{Examples} \\
\midrule
Structural & Brackets/grouping & 113 & 94.2 & $(\cdot)$, $[\cdot]$, $\{\cdot\}$ \\
Structural & Fractions/divisions & 49 & 40.8 & $\frac{a}{b}$, $a/b$ \\
Structural & Superscripts/exponents & 50 & 41.7 & $x^2$, $e^{-x}$ \\
Structural & Subscripts/indices & 74 & 61.7 & $x_i$, $a_{ij}$ \\
\midrule
Operator & Basic operators & 100 & 83.3 & $+$, $-$, $=$, $\times$, $\cdot$ \\
Operator & Integrals & 4 & 3.3 & $\int f(x)\,dx$ \\
Operator & Summations & 7 & 5.8 & $\sum_i x_i$ \\
Operator & Optimization operators & 2 & 1.7 & $\min$, $\max$ \\
\midrule
Symbol & Greek letters & 35 & 29.2 & $\alpha$, $\beta$, $\lambda$ \\
Symbol & Latin letters & 111 & 92.5 & $x$, $y$, $A$ \\
Symbol & Capitalization & 85 & 70.8 & $E$, $X$, $P$ \\
\midrule
Lexical & Variable names & 46 & 38.3 & GDP, ROI, loss \\
\midrule
Formatting & Punctuation/separators & 52 & 43.3 & $,$, $;$, $:$, $\mid$ \\
\bottomrule
\end{tabular}
\end{table}

\begin{table*}[t]
\centering
\caption{BLEU scores (\%) for formulas containing specific notation features.}
\label{tab:feature_bleu}
\setlength{\tabcolsep}{3.5pt}
\footnotesize
\begin{tabular}{llrr|cc|cc}
\toprule
& & & & \multicolumn{2}{c|}{\textbf{Simple transcription}} & \multicolumn{2}{c}{\textbf{Extended transcription}} \\
\cmidrule(lr){5-6}\cmidrule(lr){7-8}
\textbf{Category} & \textbf{Feature} & \textbf{$n$} & \textbf{Cov.} & \textbf{Without text. context} & \textbf{With text. context} & \textbf{Without text. context} & \textbf{With text. context} \\
\midrule
All & All formulas & 120 & 100.0 & \textbf{69.3} & 57.9 & 29.0 & \textbf{40.4} \\
\midrule
Structural & Brackets/grouping & 113 & 94.2 & \textbf{67.7} & 57.5 & 29.8 & \textbf{40.3} \\
Structural & Fractions/divisions & 49 & 40.8 & \textbf{68.4} & 57.6 & 33.1 & \textbf{41.7} \\
Structural & Superscripts/exponents & 50 & 41.7 & \textbf{67.3} & 56.6 & 33.5 & \textbf{43.1} \\
Structural & Subscripts/indices & 74 & 61.7 & \textbf{66.0} & 55.5 & 31.5 & \textbf{42.2} \\
\midrule
Operator & Basic operators & 100 & 83.3 & \textbf{67.3} & 57.9 & 31.7 & \textbf{42.0} \\
Operator & Integrals & 4 & 3.3 & \textbf{63.5} & 54.9 & 34.0 & \textbf{42.1} \\
Operator & Summations & 7 & 5.8 & \textbf{78.2} & 67.6 & 35.3 & \textbf{44.1} \\
Operator & Optimization operators & 2 & 1.7 & \textbf{59.5} & 55.4 & 39.8 & \textbf{44.4} \\
\midrule
Symbol & Greek letters & 35 & 29.2 & \textbf{68.1} & 54.0 & 27.6 & \textbf{39.1} \\
Symbol & Latin letters & 111 & 92.5 & \textbf{67.8} & 56.9 & 29.7 & \textbf{40.5} \\
Symbol & Capitalization & 85 & 70.8 & \textbf{65.8} & 54.7 & 29.6 & \textbf{39.7} \\
\midrule
Lexical & Variable names & 46 & 38.3 & \textbf{65.2} & 54.7 & 30.3 & \textbf{38.0} \\
\midrule
Formatting & Punctuation/separators & 52 & 43.3 & \textbf{66.5} & 55.7 & 32.8 & \textbf{42.9} \\
\bottomrule
\end{tabular}
\end{table*}

\section{Experiments and Results}
\label{sec:results}

This section reports the evaluation results and analyzes performance differences across domains, formula complexity levels, notation features, and context settings.

\subsection{Overall Performance Across Domains}

Tables~\ref{tab:bleu_domain_full} and \ref{tab:bleu_complexity} report mean formula-level BLEU (\%) scores across domains and symbol-count ranges for the five evaluated LLMs: Gemini 2.5 Flash (Gem), DeepSeek V3.2 (Deep), GPT-5.2 (GPT), Claude Sonnet 4.5 (Clau), and Grok 4.1 Fast (Gr). Scores are presented for \textit{simple} and \textit{extended} transcriptions under \textit{context-free} and \textit{context-aware} settings. Bold values indicate the highest BLEU score within each domain or symbol-count range and setting.

Table~\ref{tab:significance} reports paired $t$-tests between the LLM with the highest mean formula-level BLEU in each setting and the remaining LLMs. Asterisks indicate significant differences after Holm correction at $\alpha = 0.05$.

For \textit{simple} transcriptions without textual context, GPT-5.2 achieved the highest mean formula-level BLEU (75.9\%), followed by Grok 4.1 Fast (73.3\%) and Claude Sonnet 4.5 (71.3\%). Paired $t$-tests showed that GPT-5.2 significantly outperformed Gemini 2.5 Flash and DeepSeek V3.2, but not Grok 4.1 Fast or Claude Sonnet 4.5.

Providing textual context reduced BLEU for \textit{simple} transcriptions across all LLMs. GPT-5.2 decreased from 75.9\% to 64.9\% (-14.5\% relative), Grok 4.1 Fast from 73.3\% to 60.2\% (-17.9\%), and Claude Sonnet 4.5 from 71.3\% to 61.1\% (-14.3\%). This suggests that concise structural readings already closely matched the references, while textual context introduced lexical variation that reduced n-gram overlap.

\textit{Extended} transcriptions exhibited lower formula-level BLEU values due to greater lexical diversity. Without context, scores ranged from 21.0\% (GPT-5.2) and 22.7\% (DeepSeek V3.2) to 36.5\% (Claude Sonnet 4.5). With context, performance improved consistently: GPT-5.2 increased from 21.0\% to 42.7\% (+103.3\% relative), Grok 4.1 Fast from 35.6\% to 48.8\% (+37.1\%), Claude Sonnet 4.5 from 36.5\% to 45.2\% (+23.8\%), and Gemini 2.5 Flash from 29.1\% to 31.8\% (+9.3\%). Paired $t$-tests found no significant difference between the two highest-scoring LLMs, Claude Sonnet 4.5 and Grok 4.1 Fast, in either context setting.

BLEU scores varied across scientific domains: For \textit{simple} transcriptions in the \textit{context-free} setting, GPT-5.2 reached 80.8\% in \textit{Physics} and 77.5\% in \textit{Chemistry}, while Claude Sonnet 4.5 reached 80.2\% in \textit{Statistics} and 79.0\% in \textit{Economics}. For \textit{extended} transcriptions in the \textit{context-aware} setting, Grok 4.1 Fast led in \textit{Statistics} (59.2\%) and \textit{Mathematics} (54.4\%), and Claude Sonnet 4.5 in \textit{Physics} (44.8\%). Since no domain-level significance tests were conducted, these comparisons are descriptive and may reflect differences in notation conventions and symbol usage across disciplines.

\subsection{Impact of Formula Complexity}

Performance across symbol-count ranges was non-monotonic rather than declining uniformly with formula complexity. For \textit{simple} transcriptions in the \textit{context-free} setting, average BLEU decreased from 80.2\% for 1--10 symbols to 61.9\% for 11--20, rose to 71.4\% for 21--30, and fell to 63.8\% for 31--40 symbols. The \textit{context-aware} setting showed a similar pattern. Spearman's rank correlations based on individual symbol counts assessed the overall relationship between formula length and BLEU. For \textit{simple} transcriptions without context, BLEU decreased significantly as formula length increased ($\rho=-.237$, $p_{\mathrm{Holm}}=.018$). With context, the relationship was weaker and not significant ($\rho=-.095$, $p_{\mathrm{Holm}}=.304$). Thus, the higher BLEU for the 21--30 symbol range does not indicate a reversal of the overall trend, since each range contains only 30 formulas and its average depends on which formulas it contains. For \textit{extended} transcriptions, BLEU tended to increase with formula length, and this relationship was significant both without context ($\rho=.515$, $p_{\mathrm{Holm}}<.004$) and with context ($\rho=.259$, $p_{\mathrm{Holm}}=.012$).

\begin{table*}[h!]
\centering
\caption{Five most frequent notation features involved in token-level confusion pairs (\%).}
\label{tab:feature_errors}
\setlength{\tabcolsep}{6pt}
\footnotesize
\begin{tabular}{ll|cc|cc}
\toprule
& &
\multicolumn{2}{c|}{\textbf{Simple transcription}} &
\multicolumn{2}{c}{\textbf{Extended transcription}} \\
\cmidrule(lr){3-4}\cmidrule(lr){5-6}
\textbf{Category} &
\textbf{Feature} &
\textbf{Without textual context} &
\textbf{With textual context} &
\textbf{Without textual context} &
\textbf{With textual context} \\
\midrule
Structural & \textbf{Brackets/grouping}      & 37.9 & 32.8 & 23.8 & 23.9 \\
Operator   & \textbf{Basic operators}        &  5.3 &  7.7 & 19.1 & 19.1 \\
Formatting & \textbf{Punctuation/separators} & 20.7 & 13.2 &  8.6 &  9.9 \\
Symbol     & \textbf{Latin letters}          &  7.5 &  7.2 & 11.9 & 13.4 \\
Structural & \textbf{Fractions/divisions}    &  6.9 & 16.0 &  7.9 &  9.3 \\
\bottomrule
\end{tabular}
\end{table*}

\subsection{Notation-Level Analysis}

Tables~\ref{tab:notation_coverage} and \ref{tab:feature_bleu} provide a more detailed analysis of notation-specific behavior. Our corpus covers frequent structural features such as \textit{brackets/grouping} (94.2\%), \textit{subscripts/indices} (61.7\%), \textit{fractions/divisions} (40.8\%), and \textit{superscripts/exponents} (41.7\%). Formulas containing structural features such as \textit{fractions/divisions}, \textit{superscripts/exponents}, and \textit{summations} generally achieved comparatively high BLEU scores, while lower scores were observed for \textit{variable names}, \textit{capitalization}, and \textit{Greek letters}, suggesting greater challenges for lexical choices and symbol interpretation. For \textit{extended} transcriptions, context improved performance for all analyzed features, with the largest relative improvements for \textit{Greek letters} (+41.7\%), \textit{capitalization} (+34.1\%), and \textit{variable names} (+25.4\%). Since notation features frequently co-occur, these scores indicate associations rather than isolated feature effects. Results for rare features such as \textit{integrals} and \textit{optimization operators} should be interpreted cautiously.

Table~\ref{tab:feature_errors} complements the formula-level BLEU analysis by identifying the notation features most frequently involved in token-level confusion pairs. The reported percentages indicate the share of all confusion-pair occurrences assigned to each notation feature within the respective transcription setting. The five features shown in the table were selected according to their total number of occurrences across all four settings.

Across all four settings, \textit{brackets/grouping} accounts for the largest share of token-level confusion pairs (23.8--37.9\%). Frequent aligned pairs in this category include ``open parenthesis'' $\leftrightarrow$ ``the quantity'', ``open square bracket'' $\leftrightarrow$ ``the concentration of'', and ``open curly brace'' $\leftrightarrow$ ``the set of'', where $\leftrightarrow$ denotes aligned tokens in the reference and generated transcriptions. For example, ``open parenthesis'' $\leftrightarrow$ ``the quantity'' indicates that the reference explicitly verbalized the opening parenthesis, whereas the generated transcription introduced the expression inside it with ``the quantity''. These expressions are therefore not semantic equivalents but different verbalizations of the same part of the formula. Such cases show that LLMs may describe expressions inside grouping symbols without explicitly naming the symbols. Omitting opening or closing grouping markers can make reconstructing the original notation more difficult. Other frequent confusion pairs include ``comma'' $\leftrightarrow$ ``and'', ``times'' $\leftrightarrow$ ``by'', ``equals'' $\leftrightarrow$ ``to'', and ``over'' $\leftrightarrow$ ``by''. These pairs show that many remaining differences arise from alternative lexical choices in mathematical verbalization rather than incorrect structural interpretation.

The distribution of confusion pairs differs substantially between transcription types: For \textit{simple} transcriptions without textual context, \textit{punctuation/separators} account for 20.7\% of all confusion-pair occurrences, with ``comma'' $\leftrightarrow$ ``and'' being the most frequent aligned pair. Their share decreases to 13.2\% when textual context is provided. For \textit{extended} transcriptions, \textit{basic operators} account for 19.1\% of all confusion-pair occurrences both with and without textual context. Frequent operator-related pairs include ``times'' $\leftrightarrow$ ``by'', ``times'' $\leftrightarrow$ ``and'', and ``equals'' $\leftrightarrow$ ``to'', indicating greater lexical variation in \textit{extended} transcriptions than in \textit{simple} transcriptions.

Textual context also changes the distribution of confusion pairs within \textit{simple} transcriptions. The share assigned to \textit{fractions/divisions} increases from 6.9\% without textual context to 16.0\% with textual context. Frequent pairs in this category include ``divided'' $\leftrightarrow$ ``over'', ``over'' $\leftrightarrow$ ``by'', and ``and'' $\leftrightarrow$ ``over''. At the same time, the share assigned to \textit{punctuation/separators} decreases from 20.7\% to 13.2\%. These results do not imply that textual context uniformly increases or decreases transcription differences. Instead, they show that textual context changes which notation features contribute most strongly to the observed token-level differences. 

\subsection{Scope and Limitations}

The controlled design of our corpus enables systematic comparisons across six scientific domains, four symbol-count ranges, two transcription types, and two context settings. The study is limited to 120 English-language formulas from Wikipedia-derived documents, so future work should examine other document types, languages, and broader formula distributions. Multi-reference BLEU accounts for multiple valid transcriptions but measures reference overlap rather than semantic correctness or listener reconstructability. The same four fixed prompt variants were used for all LLMs, sharing core instructions and differing only through condition-specific additions for transcription type and textual context. Each LLM produced one output per formula and condition, so alternative prompt wording and repeated generations were not examined. The same OpenRouter request setup was used throughout, with generation parameters unspecified and provider routing handled automatically, so results reflect the applied provider defaults. Some notation features are rare and frequently co-occur, limiting fine-grained attribution. Annotator disagreements were resolved by consensus, with no separate pre-consensus agreement metric.

\section{Conclusion and Future Work}
\label{sec:conclusion}

Our work presents a systematic multi-domain evaluation of five state-of-the-art LLMs for mathematical formula verbalization across context settings, scientific domains, and formula complexity levels. The evaluated LLMs achieved high reference overlap for concise structural readings, with GPT-5.2 reaching the highest mean formula-level BLEU of 75.9\% for \textit{simple} transcriptions without context. This highlights the potential of LLMs for notation-oriented formula verbalization.

For \textit{simple} transcriptions, textual context increased lexical variation and lowered BLEU scores. In contrast, \textit{extended} transcriptions achieved higher BLEU with textual context, suggesting that LLMs can use surrounding context when generating more explanatory descriptions.

Performance varied across scientific domains, formula complexity levels, and notation features. Formulas with \textit{variable names}, \textit{capitalization}, and \textit{Greek letters} tended to show lower BLEU than those with structural features such as fractions, superscripts, and summations. Token-level confusion analysis further showed that grouping symbols were most frequently involved in transcription differences across all settings, while other confusion pairs varied between \textit{simple} and \textit{extended} transcriptions and with textual context. Overall, LLMs capture many structural aspects of mathematical notation, while semantic interpretation and explicit preservation of hierarchical grouping remain key challenges.

Future work should complement BLEU-based evaluation with human intelligibility and reconstruction studies assessing whether listeners can recover the original formulas from generated speech. Further directions include reasoning-oriented prompting, ensemble methods, larger multi-domain corpora, and closer integration with text-to-speech systems.

\section*{Acknowledgments}

This research was funded by IU Int. University of Applied Sciences (\textit{IU Incubator}) from April 2026 to March 2028.

\bibliographystyle{IEEEtran}
\bibliography{references.bib}

\end{document}